\documentclass{acmsiggraph}

\TOGonlineid{45678}

\usepackage{lipsum}

\newcommand\blfootnote[1]{%
  \begingroup
  \renewcommand\thefootnote{}\footnote{#1}%
  \addtocounter{footnote}{-1}%
  \endgroup
}

\newcommand{\edit}[1]{{#1}}

\title{Learning-Based View Synthesis for Light Field Cameras}

\author{Nima Khademi Kalantari$^{1}$  \hspace{0.2in} Ting-Chun Wang$^{2}$ \hspace{0.2in} Ravi Ramamoorthi$^{1}$ \vspace{0.1in}
\\ $^{1}$University of California, San Diego \hspace{0.4in} $^{2}$University of California, Berkeley}
\pdfauthor{Nima Khademi Kalantari, Ting-Chun Wang, Ravi Ramamoorthi}

\keywords{view synthesis, light field, convolutional neural network, disparity estimation}

\begin{document}

%\copyrightyear{2016}
%\setcopyright{acmlicensed}
%\conferenceinfo{SA '16 Technical Papers,}{December 05 - 08, 2016, , Macao}
%\isbn{978-1-4503-4514-9/16/12}
%\doi{http://dx.doi.org/10.1145/2980179.2980251}

\maketitle

\begin{abstract}
With the introduction of consumer light field cameras, light field imaging has recently become widespread. However, there is an inherent trade-off between the angular and spatial resolution, and thus, these cameras often sparsely sample in either spatial or angular domain. In this paper, we use machine learning to mitigate this trade-off. Specifically, we propose a novel learning-based approach to synthesize new views from a sparse set of input views. We build upon existing view synthesis techniques and break down the process into disparity and color estimation components. We use two sequential convolutional neural networks to model these two components and train both networks simultaneously by minimizing the error between the synthesized and ground truth images. We show the performance of our approach using only four corner sub-aperture views from the light fields captured by the Lytro Illum camera. Experimental results show that our approach synthesizes high-quality images that are superior to the state-of-the-art techniques on a variety of challenging real-world scenes. We believe our method could potentially decrease the required angular resolution of consumer light field cameras, which allows their spatial resolution to increase.

\end{abstract}

\begin{CCSXML}
<ccs2012>
<concept>
<concept_id>10010147.10010371.10010382</concept_id>
<concept_desc>Computing methodologies~Image manipulation</concept_desc>
<concept_significance>500</concept_significance>
</concept>
<concept>
<concept_id>10010147.10010371.10010382.10010236</concept_id>
<concept_desc>Computing methodologies~Computational photography</concept_desc>
<concept_significance>300</concept_significance>
</concept>
</ccs2012>
\end{CCSXML}

\ccsdesc[500]{Computing methodologies~Image manipulation}
\ccsdesc[300]{Computing methodologies~Computational photography}


% The next three commands are required, and insert the user-generated keywords, 
% The CCS concepts list, and the rights management text.
% Please make sure there is a blank line between each of these three commands.

\keywordlist

\conceptlist

\blfootnote{The definitive Version of Record was published in ACM TOG and can be accessed via http://dx.doi.org/10.1145/2980179.2980251.}
\blfootnote{}
\blfootnote{}
\blfootnote{}
\blfootnote{}
\blfootnote{}
\blfootnote{}
\blfootnote{}
\blfootnote{}
\blfootnote{}

%\printcopyright

\section{Introduction}
\label{sec:Introduction}

Light fields provide a rich representation of real-world scenes, enabling exciting applications such as refocusing and viewpoint change. Generally, they are obtained by capturing a set of 2D images from different views~\cite{Levoy96,Wilburn05} or using a microlens array~\cite{Adelson92,Ng2005,Georgeiv06}. The early light field cameras required custom-made camera setups which were bulky and expensive, and thus, not available to the general public. Recently, there has been renewed interest in light field imaging with the introduction of commercial light field cameras such as Lytro~\shortcite{Lytro} and RayTrix~\shortcite{RayTrix}. However, because of the limited resolution of the sensors, there is an inherent trade-off between angular and spatial resolution, which means the light field cameras sample sparsely in either the angular or spatial domain. For example, Pelican cameras~\cite{Pelican} have an array of $2\times 2$ cameras. 

To mitigate this problem, we propose a learning-based approach to synthesize novel views from a sparse set of input views captured using consumer light field cameras. Inspired by the recent success of deep learning in a variety of applications, such as image denoising~\cite{Burger12}, super-resolution~\cite{Dong14}, and deblurring~\cite{Sun15}, we propose to use convolutional neural networks (CNN) to predict novel views using the sparse input views and the position of the novel view in the light field. However, the major challenge is that training a single end-to-end CNN for this task is difficult, producing novel views that are quite blurry, as shown in Fig.~\ref{fig:Naive}.

\begin{figure}[t]
\centering
\ifx\ShowTempImages\undefined
   \includegraphics[width=1\linewidth]{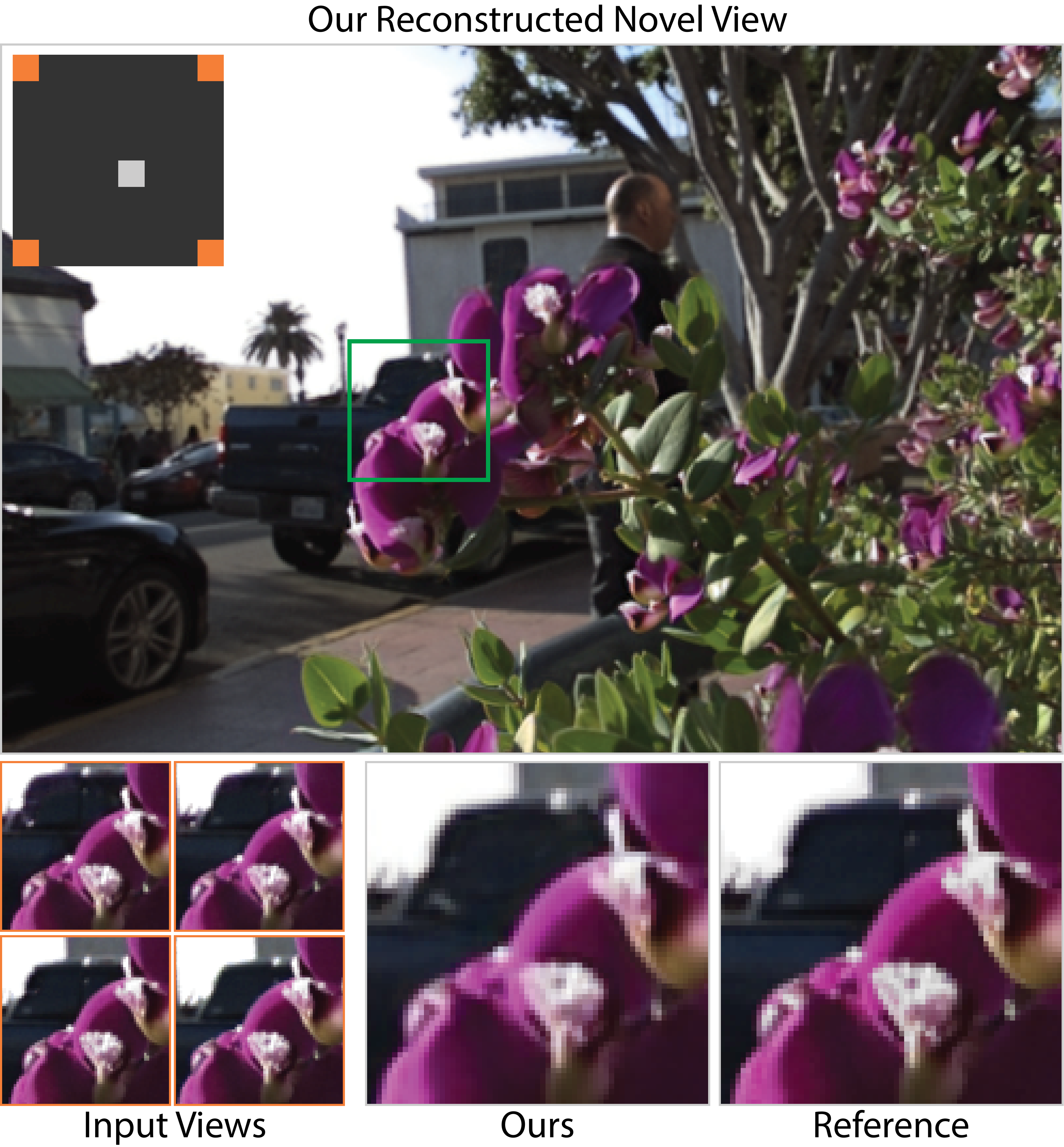}
\else
   \includegraphics[width=1\linewidth]{Images/Teaser_PlaceHolder}
\fi
\vspace{-0.2in}
\caption{\label{fig:Teaser}
We propose a learning-based approach to synthesize novel views from a sparse set of input views captured with a consumer light field camera. We capture a light field with angular resolution of $8\times 8$ using a Lytro Illum camera and use only the four corner sub-aperture images as our input. Our learning-based method is able to handle the occlusion boundaries between the flower and the background and produce a plausible image which is comparable to the ground truth image. We show comparison against state-of-the-art approaches in Fig.~\ref{fig:MainComparison}. The figures throughout the paper are best seen by zooming in to the electronic version of the paper.}
\vspace{-0.2in}
\end{figure}

\begin{figure*}[t]
\centering
\ifx\ShowTempImages\undefined
	\includegraphics[width=\linewidth]{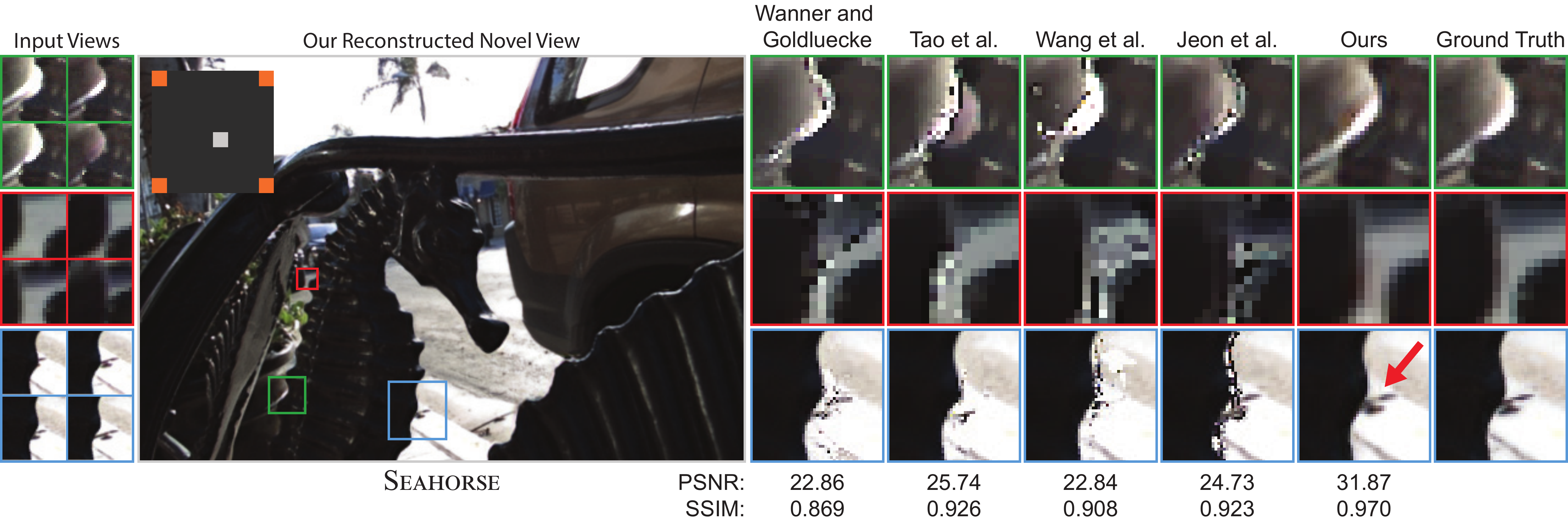}
\else
	\includegraphics[width=\linewidth]{Images/Motivation_PlaceHolder}
\fi
\vspace{-0.25in}
\caption{\label{fig:Motivation}
This scene demonstrates a seahorse statue in front of a car (on the right) and a street. The method of Wanner and Goldluecke~[2014] takes the estimated disparity at the input views and synthesizes images at novel views. We compare our approach against this method using several state-of-the-art light field disparity estimation methods as its input. We use only four corner images of a light field captured with a Lytro Illum camera as the input and synthesize one of the middle views with all these techniques. Although these disparity estimation methods produce reasonable disparity maps, they are not specifically designed for view synthesis. Furthermore, Wanner and Goldluecke's method assumes the images to be ideal, while, in practice, the images captured with commercial light field cameras often contain noise and suffer from optical distortions. Therefore, these methods produce results with tearing, ghosting, and other artifacts. Our method, however, learns to handle these inaccuracies and produces disparities that are suitable for this application. As a result, we generate high-quality images without disturbing artifacts that are superior to other methods, both visually and numerically in terms of PSNR (dB) and structural similarity (SSIM). Larger SSIM values show better perceptual quality. Note that only our method is able to reconstruct the highlight and the dark structure in the green and blue insets, respectively. See Fig.~\ref{fig:DepthComp} for comparison of the estimated disparities.}
\vspace{-0.25in}
\end{figure*}

Existing view synthesis approaches~\cite{Chaurasia11,Chaurasia13,Wanner14} typically first estimate the depth at the input views and use it to warp the input images to the novel view. They then combine these images in a specific way (e.g., by weighting each warped image~\cite{Chaurasia13}) to obtain the final novel view image. To make the learning more tractable, we build upon these methods and break down the task into disparity\footnote{Because of the regularity of camera positions in structured light fields, depth and disparity are closely related and we use them interchangeably.} and color estimation components. The main contribution of our work is to use machine learning to model these two components and train both models by directly minimizing the error between the synthesized and ground truth images. In our system, we use two sequential CNNs for estimating the disparity and the final pixel colors. Since our disparity estimation CNN is trained to directly minimize the synthesis error, our estimated disparities are more suitable for this application than existing disparity estimation techniques~\cite{Wanner12,Tao13,Wang15,Jeon15} (see Fig.~\ref{fig:DepthComp}). Moreover, since we train our system on the light fields generated by consumer light field cameras, it learns to model the noise and other inaccuracies of these cameras. Therefore, our method produces better results than the state-of-the-art optimization-based approach of Wanner and Goldluecke~\shortcite{Wanner14}, as shown in Fig.~\ref{fig:Motivation}.

We demonstrate the performance of our approach using only the four corner sub-aperture views from $8 \times 8$\footnote{The actual angular resolution of the Lytro Illum cameras is $14 \times 14$. However, the three views from each side are usually black, and thus, we use only the eight middle views in our implementation.} light fields captured by the Lytro Illum camera (see Fig.~\ref{fig:Teaser}). Experimental results demonstrate that our method outperforms state-of-the-art schemes on challenging cases. Our method is two orders of magnitude faster than the recent learning-based DeepStereo method of Flynn et al.~\shortcite{Flynn15}, taking only 12.3 seconds to synthesize an image from four input views of size $541 \times 376$. Our system could potentially be used to decrease the required angular resolution of current cameras, which allows their spatial resolution to increase. Another application of our approach is to increase the baseline of current cameras and use our method on a subset of four angular views to synthesize the in between views. In summary, we make the following contributions:

\begin{itemize}

\item We present the first machine learning approach for view synthesis using consumer light field cameras. Our system consists of disparity and color estimation components which we model using two sequential CNNs. \edit{Note that although CNNs have been recently used for light field super-resolution~\cite{Yoon15} and depth estimation~\cite{Heber16}, these methods are not able to directly synthesize novel views at arbitrary locations.} 

\item The output of our first network is disparity and typically we would need ground truth disparities to train this network. However, we show how to train both networks simultaneously by directly minimizing the error between the synthesized and ground truth images.

\item Since we train our disparity estimator in this way, our disparities are suitable for the view synthesis application. To the best of our knowledge, our method is the first to propose a disparity estimator which is specifically designed for this application.

\end{itemize}

\section{Related Work}
\label{sec:RelatedWork}

The problem of the light field's limited resolution has been extensively studied in the past and several powerful methods for increasing the resolution in both angular~\cite{Levin10,Shi14,Wanner14} and spatial~\cite{Bishop09,Cho13} domains have been proposed. For brevity, we only focus on the approaches that are designed for angular super-resolution. We start by reviewing the algorithms that specifically work for light fields and then explain the approaches that perform view synthesis for general scenes and objects.

\noindent{\bf Light Field Super-resolution} -- Levin and Durand~\shortcite{Levin10} use a prior based on the dimensionality gap to reconstruct the full 4D light field from a 3D focal stack sequence. Shi et al.~\shortcite{Shi14} leverage sparsity in the continuous Fourier spectrum to reconstruct a dense light field from a 1D set of view points. \edit{Schedl et al.~\shortcite{Schedl15} reconstruct a full light field using multidimensional patches from a sparse set of input views.} These methods require the input samples to be captured with a specific pattern and are not able to synthesize novel views at arbitrary positions. Marwah et al.~\shortcite{Marwah13} propose a dictionary-based approach to reconstruct light fields from a coded 2D projection. However, their method requires the light fields to be captured in a compressive way.  

Mitra and Veeraraghavan~\shortcite{Mitra12} introduce a patch-based approach where they model the light field patches using a Gaussian mixture model. However, this method is not robust against noise, and struggles on low-quality images taken with commercial light field cameras. Zhang et al.~\shortcite{Zhang15} propose a phase-based approach to reconstruct light fields. However, their method is limited since it is designed for a micro-baseline stereo pair. Moreover, their approach is iterative, which is often slow and prevents its usage in practice. Yoon et al.~\shortcite{Yoon15} perform spatial and angular super-resolution on light fields using convolutional neural networks (CNN). However, their method can only increase the resolution by a factor of two, and is not able to synthesize views at arbitrary locations. \edit{Layered patch-based synthesis has been proposed by Zhang et al.~\shortcite{Zhang16} for various light field editing applications. Although they show impressive results for applications like hole-filling and reshuffling, their approach has limited performance for view synthesis and is not able to handle challenging cases as shown in Fig.~\ref{fig:PlenoPatch}.}

Recently, Wanner and Goldluecke~\shortcite{Wanner14} proposed an optimization approach to reconstruct images at novel views from an input light field. Given the depth estimates at the input views, they reconstruct novel views by minimizing an objective function which maximizes the quality of the final results. Although their method produces reasonable results on dense light fields, for sparse input views, it often produces results with tearing, ghosting, and other artifacts as shown in Fig.~\ref{fig:Motivation}. We believe this is because of two main reasons. First, they estimate the disparity at the input views as a preprocess, independently of the view synthesis process. However, even state-of-the-art light field disparity estimation techniques~\cite{Wang15,Jeon15} are not typically designed to maximize the quality of synthesized views, and thus, they are not suitable for this application. Second, Wanner and Goldluecke's method assumes that the images are captured under ideal conditions. However, in practice, the images from consumer light field cameras are usually noisy and suffer from optical distortions.

\noindent{\bf View Synthesis for Scenes} -- View synthesis has a long history in both vision and graphics. One category of approaches~\cite{Eisemann08,Goesele10,Chaurasia11,Chaurasia13} synthesizes novel views of a scene in a two-step process. These methods first estimate the depth at the input views and use the depth to warp the input images to the novel view. They then produce the final image by combining these warped images. These approaches typically use multi-view stereo algorithms (e.g., PMVS by Furukawa et al.~\shortcite{Furukawa10}) to estimate depth and are not suitable for light fields with a narrow baseline. In our system, we also have depth and color estimation components. However, unlike these approaches, we use machine learning to model these two components. Furthermore, inspired by Fitzgibbon et al.'s approach~\shortcite{Fitzgibbon03}, we train both our disparity and color estimation models by directly minimizing the appearance error.

Another common approach is to synthesize images without explicitly estimating the geometry. For example, Mahajan et al.~\shortcite{Mahajan09} propose to move the gradients in the input images along a specific path to reconstruct the image at a novel view. Shechtman et al.~\shortcite{Shechtman10} propose a patch-based optimization framework to reconstruct images at novel views. However, these approaches are not able to utilize all the information available in light fields since they work on only two input images.

{\bf DeepStereo} -- Flynn et al.~\shortcite{Flynn15} has recently proposed a deep learning method to perform view synthesis on a sequence of images with wide baselines. They first project the input images on multiple depth planes. They then estimate the pixel color and weight of the image at each depth plane from these projected images. Finally, they compute a weighted average of the estimated pixel colors to obtain the final pixel color. Comparing to this approach, our system has several key differences. First, our method is specifically designed for light fields, which have much narrower baselines and more regular camera positions. Second, unlike their approach, our system explicitly estimates the disparity which could potentially be used in other applications. Finally, our system is significantly faster than their method (several minutes vs. seconds). This shows the efficiency of our system, validating more practical usage.

\noindent{\bf View Synthesis for Objects} -- Since the recent release of large datasets of 3D shape models, synthesizing object views from a single image has become popular. Kholgade et al.~\shortcite{Kholgade14} transfer texture from the corresponding 3D model to render novel views of an object. However, manual annotation is required to specify the corresponding 3D model and its placement in the image. Su et al.~\shortcite{Su14} resolve this limitation by selecting several similar models in the dataset and then interpolating between them. However, these methods heavily rely on the retrieval process and become vulnerable when a similar model cannot be found.

Recently, several algorithms have approached this problem by utilizing deep learning. Dosovitskiy et al.~\shortcite{Dosovitskiy15} train a CNN which can render images of chairs once a graphics code containing the rendering details is given. Yang et al.~\shortcite{Yang15} expand this work and decode the implicit rendering information from the input image instead of representing it explicitly as the graphics code. They then apply the desired transformation and render the new view. Tatarchenko et al.~\shortcite{Tatarchenko15} also adopt a similar approach, but do not explicitly decouple the identity and the pose. \edit{Zhou et al.~\shortcite{Zhou16} train a CNN to estimate appearance flow which is then used to warp the input image to the novel view.} These methods are specifically designed to work on objects and do not work well on general scenes. Furthermore, they only use a single image, and thus, are not able to utilize all the images in light fields.
\section{Proposed Learning-Based Algorithm}
\label{sec:Algorithm}

Given a sparse set of input views $L_{p_1}, \cdots, L_{p_N}$ and the position of the novel view $q$, our goal is to estimate the image at the novel view $L_q$. Formally, we can write this as:

\vspace{-0.1in}
\begin{equation}
\label{eq:Goal}
L_q = f(L_{p_1}, \cdots L_{p_N}, q),
\end{equation}
\vspace{-0.2in}

where $p_i$ and $q$ refer to the $(u, v)$ coordinates of the input and novel view, respectively. Here, $f$ is a function which defines the relationship between the input views and the novel view. This relationship is typically very complex as it requires finding connections between all the input views, and collecting appropriate information from each image based on the position of the novel view. Inaccuracies such as noise and optical distortions in consumer light field cameras further add to the complexity of this relationship.

Therefore, we propose to learn this relationship. Inspired by the recent success of deep learning in a variety of applications, we propose to use convolutional neural networks (CNN) as our learning model. A straightforward way to do so is to directly model the function $f$ with a CNN. In this case, the CNN takes the input views as well as the position of the novel view and outputs the image at the novel view. However, as shown in Fig.~\ref{fig:Naive}, this na\"{\i}ve solution often produces blurry results. This is mainly due to the fact that the relationship is complex and requires the network to find connections between distant pixels, which makes the training difficult.

\begin{figure*}[!ht]
\centering
\ifx\ShowTempImages\undefined
	\includegraphics[width=0.8\linewidth]{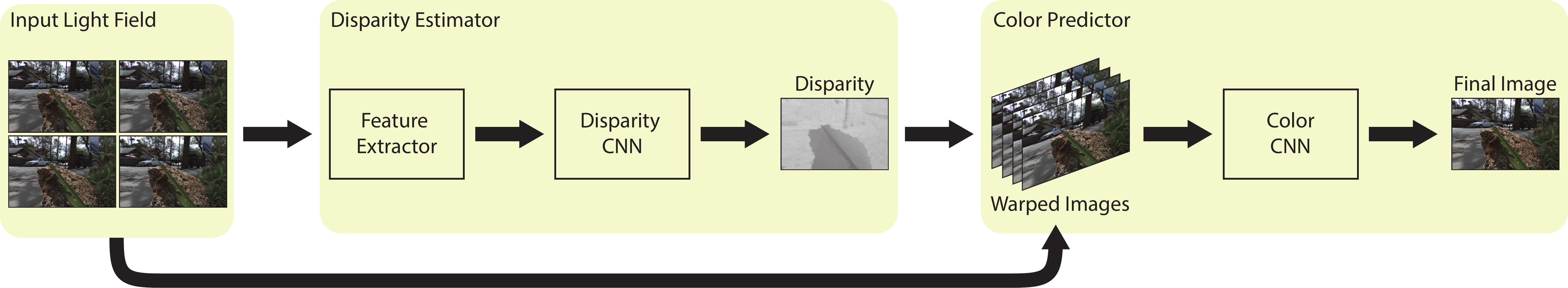}
\else
	\includegraphics[width=0.8\linewidth]{Images/Overview_PlaceHolder}
\fi
\vspace{-0.1in}
\caption{\label{fig:Overview}
Our system consists of disparity estimator and color predictor components which we model using two sequential CNNs. In our system, we first extract a set of features from the sparse input views. We then use the first CNN to estimate the disparity at the novel view. We then use this disparity to warp (backward) all the input views to the novel view. Our second CNN uses all the warped images along with a few other features to generate the final image.}
\vspace{-0.25in}
\end{figure*}

We make the training more tractable by following the pipeline of existing view synthesis techniques~\cite{Chaurasia11,Chaurasia13} and breaking the system down into disparity and color estimation components. Our main contribution is to use machine learning to model each component and train both models simultaneously by minimizing the error between the synthesized and ground truth images (see Sec.~\ref{ssec:Training}). In our system, we first estimate the disparity at the novel view from a set of features extracted from the sparse set of input views:

\vspace{-0.2in}
\begin{equation}
\label{eq:DispEst}
D_q = g_d(K),
\end{equation}
\vspace{-0.2in}

where $D_q$ is the estimated disparity at the novel view, $K$ represents a set of features including the mean and standard deviation of warped images at different disparity levels (see Sec.~\ref{ssec:DispEstimator}). Moreover, $g_d$ defines the relationship between the input features and the disparity which we model using a CNN. The estimated disparity is then used to warp the input images to the novel view. Specifically, we perform a backward warp by sampling the input images based on the disparity at the novel view (see Eq.~\ref{eq:Warping}). Finally, we estimate the image at the novel view using a set of input features including all the warped images, the estimated disparity, and the position of the novel view:

\vspace{-0.2in}
\begin{equation}
\label{eq:ColEst}
L_q = g_c(H),
\end{equation}
\vspace{-0.2in}

where $H$ represents our feature set and $g_c$ defines the relationship between these features and the final image. The overview of our system is shown in Fig.~\ref{fig:Overview}. In the next sections we describe the disparity estimator (Eq.~\ref{eq:DispEst}) and the color predictor (Eq.~\ref{eq:ColEst}) in detail.

\subsection{Disparity Estimator}
\label{ssec:DispEstimator}

The goal of this component is to estimate the disparity at the novel view $D_q$. For every pixel of the novel view image, this disparity points to the corresponding pixel in each input view:

\vspace{-0.15in}
\begin{equation}
\label{eq:Warping}
\bar{L}_{p_i}(s) = L_{p_i}\left[s + (p_i - q) \, D_q(s)\right],
\end{equation}
\vspace{-0.2in}

where $s$ is a vector containing the pixel position in the $x$ and $y$ directions. Moreover, $p_i$ and $q$ are also vectors containing the position of input and novel views in the $u$ and $v$ directions. Here, $\bar{L}_{p_i}$ is the image obtained by backward warping the input view $L_{p_i}$ using the disparity $D_q$. If the disparity is accurate, it will point to the correct pixel in the input images, and thus, all the warped images would have the same color at each pixel. However, the disparity is not known a priori and we need to estimate it first.

To estimate the disparity, we first warp (backward) all the input images to the novel view using a set of predefined disparity levels $d_1, \cdots, d_L$ as follows:

\vspace{-0.15in}
\begin{equation}
\bar{L}^{d_l}_{p_i}(s) = L_{p_i}\left[s + (p_i - q) \, d_l\right], 
\end{equation}
\vspace{-0.2in}

where $i \in \{1, \cdots, N\}$ and $l \in \{1, \cdots, L\}$. In our implementation we use $L = 100$ disparity levels in the range $[-21, 21]$ pixels. We use ideas from the depth estimation approach of Tao et al.~\shortcite{Tao13}, which is also the core of other recent techniques~\cite{Wang15,Tao15}, to extract a set of features from these warped images. Specifically, we compute the mean and standard deviation of all the warped input images at each disparity level as follows:

\vspace{-0.25in}
\begin{align}
\label{eq:Features}
M^{d_l}(s) &= \frac{1}{N} \sum_{i = 1}^N \bar{L}^{d_l}_{p_i}(s) \\ \nonumber
V^{d_l}(s) &= \sqrt{\frac{1}{N-1} \sum_{i = 1}^N (\bar{L}^{d_l}_{p_i}(s) - M^{d_l}(s))^2}.
\end{align}
\vspace{-0.2in}

\begin{figure}[t]
\centering
\ifx\ShowTempImages\undefined
	\includegraphics[width=\linewidth]{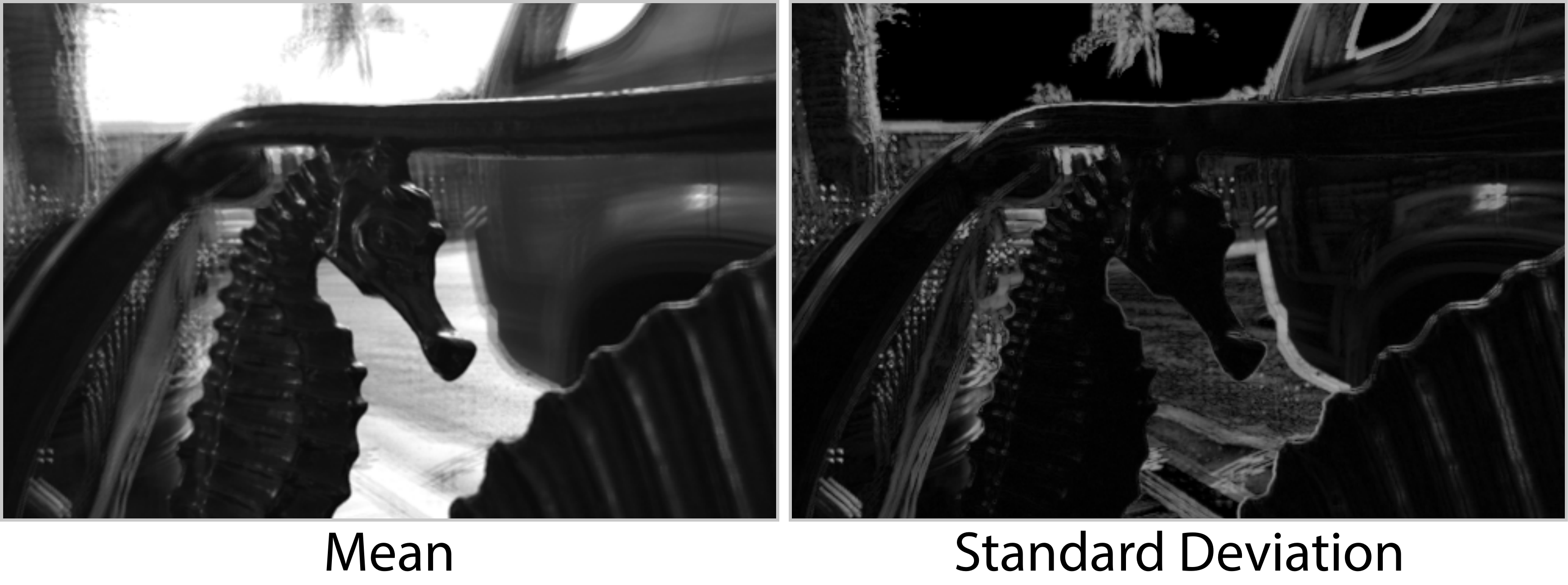}
\else
	\includegraphics[width=\linewidth]{Images/Features_PlaceHolder}
\fi
\vspace{-0.2in}
\caption{\label{fig:Features}
We show our mean and standard deviation features for a single disparity level on the {\sc Seahorse} scene. This particular disparity level is correct for the foreground. As a result, the seahorse in the mean image is sharp, while the background is blurry. Moreover, the standard deviation is small on the seahorse and is large on the background. Our network learns to use these features to estimate the correct disparity for each pixel.}
\vspace{-0.1in}
\end{figure}

We generate our input features by concatenating the mean and standard deviation for all the disparity levels $K = \{M^{d_1}, V^{d_1}, \cdots, M^{d_L}, V^{d_L}\}$ (see Fig.~\ref{fig:Features}). Since we use 100 disparity levels, our feature vector has 200 channels. 

As discussed earlier, all the warped input views have photo-consistency for the correct disparity level. Therefore, existing techniques~\cite{Tao13,Wang15} typically select the disparity level that has the minimum standard deviation and maximum mean contrast. Since the obtained disparity from this process is usually noisy, these methods use an optimization scheme to regularize the disparity. Although these approaches produce high-quality disparity maps, they are not specifically designed for the view synthesis application. Therefore, as shown in Fig.~\ref{fig:DepthComp}, they often have artifacts around the occlusion boundaries which are important regions for synthesizing high-quality images.

We avoid this problem using a learning system to estimate the optimal disparity map from the input features. As discussed in Sec.~\ref{ssec:Training}, we train our system by minimizing the error between the estimated and ground truth novel view images. Note that one may train the disparity estimator by minimizing the error between the estimated and ground truth disparities. However, we avoid this alternative since it has two main drawbacks. First, training in this way requires a database of the light fields with their corresponding ground truth disparities which is difficult to obtain. Second, if the final goal is to synthesize novel views, the disparity does not need to always be accurate. For example, a constant color region can be easily reconstructed even with inaccurate disparity.

As our model, we use a deep CNN, consisting of four convolutional layers with decreasing kernel sizes as depicted in Fig.~\ref{fig:Network}. All the layers with the exception of the last layer are followed by a rectified linear unit. Next, we explain our color predictor component.

\subsection{Color Predictor}
\label{ssec:ColorPredictor}

The goal of this component is to estimate the final color using the disparity, estimated by the first CNN. The estimated disparity can be used to simply warp the input views to the novel view using Eq.~\ref{eq:Warping}. Existing view synthesis techniques~\cite{Chaurasia11,Chaurasia13,Wanner14} have a specific way of combining these warped images and generating the final image. For example, Chaurasia et al.~\shortcite{Chaurasia13} obtain the final image by computing the weighted average of all the warped images. However, these approaches are usually simple and do not properly model the relationship between the warped and final synthesized images which is complex because of occlusion.

In contrast, we propose to learn this relationship. We estimate the final image from a set of input features including all the warped images, the estimated disparity, and the position of the novel view. Specifically, our feature vector is $H = \{\bar{L}_{p_1}, \cdots, \bar{L}_{p_N}, D_q, q\}$. Note that the disparity is useful to detect the occlusion boundaries and collect appropriate information from the warped images near these regions. Moreover, the position of the novel view can potentially be used to weight a particular image more in reconstructing the novel view. For example, if $q$ is close to $p_1$, $\bar{L}_{p_1}$ should be heavily used in reconstructing the novel view at position $q$. Although we do not explicitly model the occlusion, our system learns to reconstruct the final image by relying on the images with valid information in the occluded regions.

\begin{figure}[t]
\vspace{-0.0in}
\centering
\ifx\ShowTempImages\undefined
	\includegraphics[width=\linewidth]{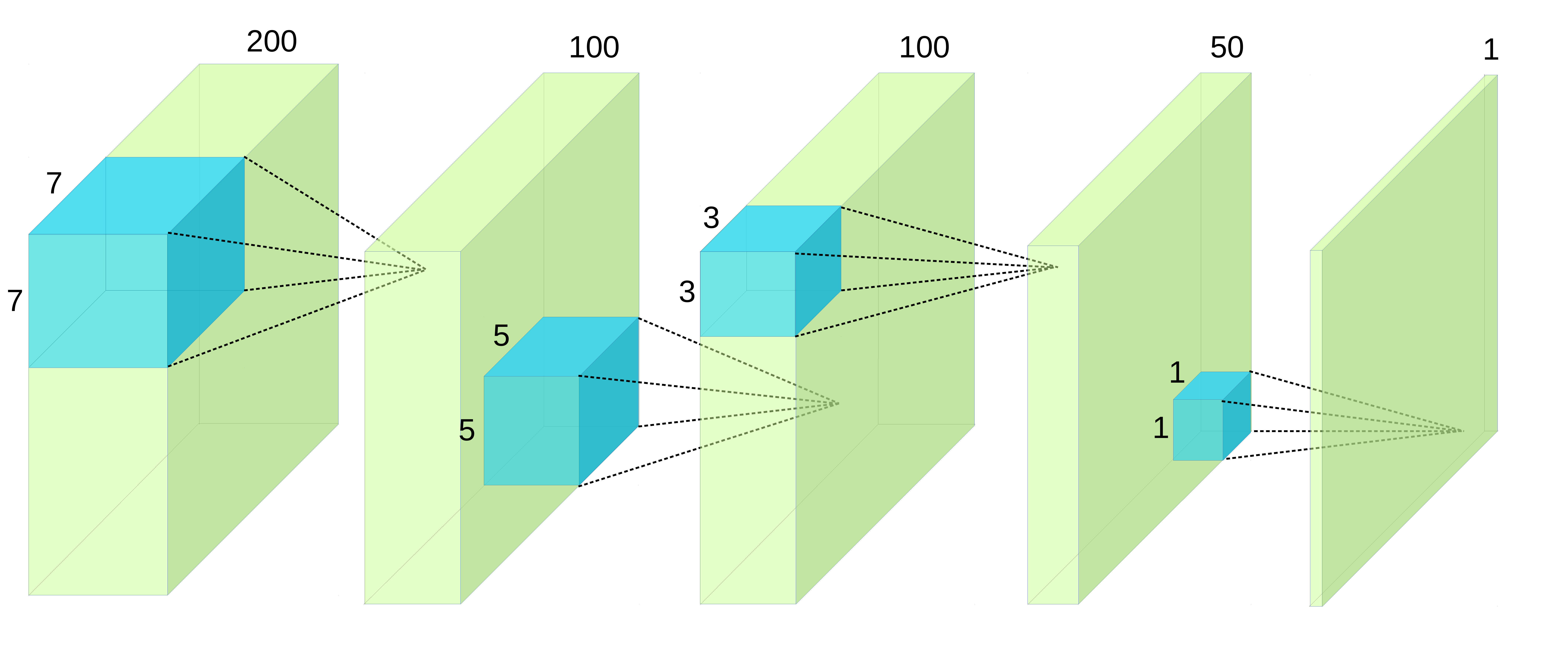}
\else
	\includegraphics[width=\linewidth]{Images/OurNetwork_PlaceHolder}
\fi
\vspace{-0.25in}
\caption{\label{fig:Network}
Our disparity estimator network consists of four convolutional layers with decreasing kernel sizes. All the layers are followed by a rectified linear unit (ReLU). Our color predictor network has a similar architecture with different number of input and output channels. We only used convolutional layers to operate on images with any size. We empirically found that this architecture properly models both our components and is reasonably fast.}
\vspace{-0.25in}
\end{figure}

Here, we use a similar deep network as in Fig.~\ref{fig:Network} with different number of inputs and outputs. In this case, our input has $3N+3$ channels and the output is an RGB image which has 3 channels. In the next section, we discuss the details of training our system.

\subsection{Training}
\label{ssec:Training}

In order to synthesize high-quality images that are close to the ground truth, we train the networks by minimizing the $\mathcal{L}_2$ distance between the synthesized and ground truth images:

\vspace{-0.2in}
\begin{equation}
\label{eq:Error}
E = \sum_{k = 1}^3(\hat{L}_{q, k} - L_{q, k})^2,
\end{equation}
\vspace{-0.2in}

where the summation is over the RGB channels, $L_{q,k}$ is the ground truth image at the novel view, and $\hat{L}_{q, k}$ is our estimated image which is obtained by Eqs.~\ref{eq:DispEst}~and~\ref{eq:ColEst}. In order to use a gradient descent based technique to minimize our energy function, we need to compute the derivative of the error in Eq.~\ref{eq:Error} with respect to both networks' weights, i.e., $\partial E/\partial w_d$ and $\partial E/\partial w_c$, where $w_d$ and $w_c$ are vectors and refer to all the weights of the disparity and color estimator networks, respectively. 

Since the color predictor network directly outputs the synthesized image, $\partial E/\partial w_c$ can be easily computed as in standard backpropagation~\cite{Rumelhart86}. For $\partial E/\partial w_d$, we use the chain rule to break down the derivative into three terms as follows:

\vspace{-0.15in}
\begin{equation}
\frac{\partial E}{\partial w_d} = \sum_{k = 1}^3\left[\frac{\partial E}{\partial \hat{L}_{q, k}} \frac{\partial \hat{L}_{q, k}}{\partial D_q}\right] \frac{\partial D_q}{\partial w_d}.
\end{equation}
\vspace{-0.15in}

Since our error is quadratic, the first term can be easily calculated. The last term is the derivative of the disparity estimation network's output with respect to its weights which can be calculated as usual~\cite{Rumelhart86}. The middle term is the derivative of the final image with respect to the estimated disparity. Note that the disparity is used to generate a set of features $H$ (see Sec.~\ref{ssec:ColorPredictor}). These features are then used by the color estimator network to produce the final image. Therefore, we have:

\vspace{-0.15in}
\begin{equation}
\frac{\partial \hat{L}_{q, k}}{\partial D_q} =  \sum_{t = 1}^{3N+3} \frac{\partial \hat{L}_{q, k}}{\partial H_t} \frac{\partial H_t}{\partial D_q},
\end{equation}
\vspace{-0.15in}

where the summation is over the individual channels of the feature vector. Here, the first term is the derivative of the color predictor network's output with respect to its input and is straightforward to calculate. For the second term, we need to investigate each channel separately. The first $3N$ channels of our input feature vector $H$ are the warped images, and thus, $\partial H_t/\partial D_q$ is basically the derivative of the warping function in Eq.~\ref{eq:Warping}. Fortunately, since we use bicubic interpolation to compute the color values, this function is differentiable. For simplicity of the implementation, we numerically calculate this gradient. The feature at the next channel $H_{3N+1}$ is the estimated depth and its derivative is equal to one. Finally, the last two channels are the position of the novel view which are independent of the disparity, and thus, their gradient is equal to zero.

At every iteration of the training, we use these gradients to update both networks' weights in the opposite direction of the gradients. We used a set of 100 light fields captured with the Lytro Illum camera in our training set. To handle a diverse test set, we ensured our training set contained a variety of different scenes including bicycles, cars, trees, and foliage (see supplementary materials). We captured most of these images ourselves, and obtained some of them from Raj et al.'s dataset~\shortcite{Raj16}. These light fields have angular resolution of $8 \times 8$ from which we only used the four corner sub-aperture images as our input. For each light field, we randomly selected four novel view positions from the original $8\times 8$ grid. For each novel view position we extracted a set of features (see Eq.~\ref{eq:Features}) and used the original captured image at that position as the ground truth image. 

Since training on the full images is slow, we extracted patches of size $60\times 60$ with a stride of 16 pixels from the full images. This resulted in over 100,000 patches which we used to train our system. Note that for every input patch, our system outputs a patch of size $36 \times 36$ (reduced size is due to convolutions). These output patches are then compared to the ground truth patches and the error at each pixel is backpropagated to train the networks. Therefore, in practice, we had more than 100,000,000 examples which we found to be sufficient to properly train both networks. We used mini-batches of size 20 to have the best trade-off between speed and convergence. We randomly initialized our networks' weights using the Xavier approach~\cite{Glorot10} and trained our system using the ADAM solver~\cite{Kingma14}, with ${\beta_1 = 0.9}$, ${\beta_2 = 0.999}$, and a learning rate of 0.0001.

\begin{figure}[t]
\centering
\includegraphics[width=\linewidth]{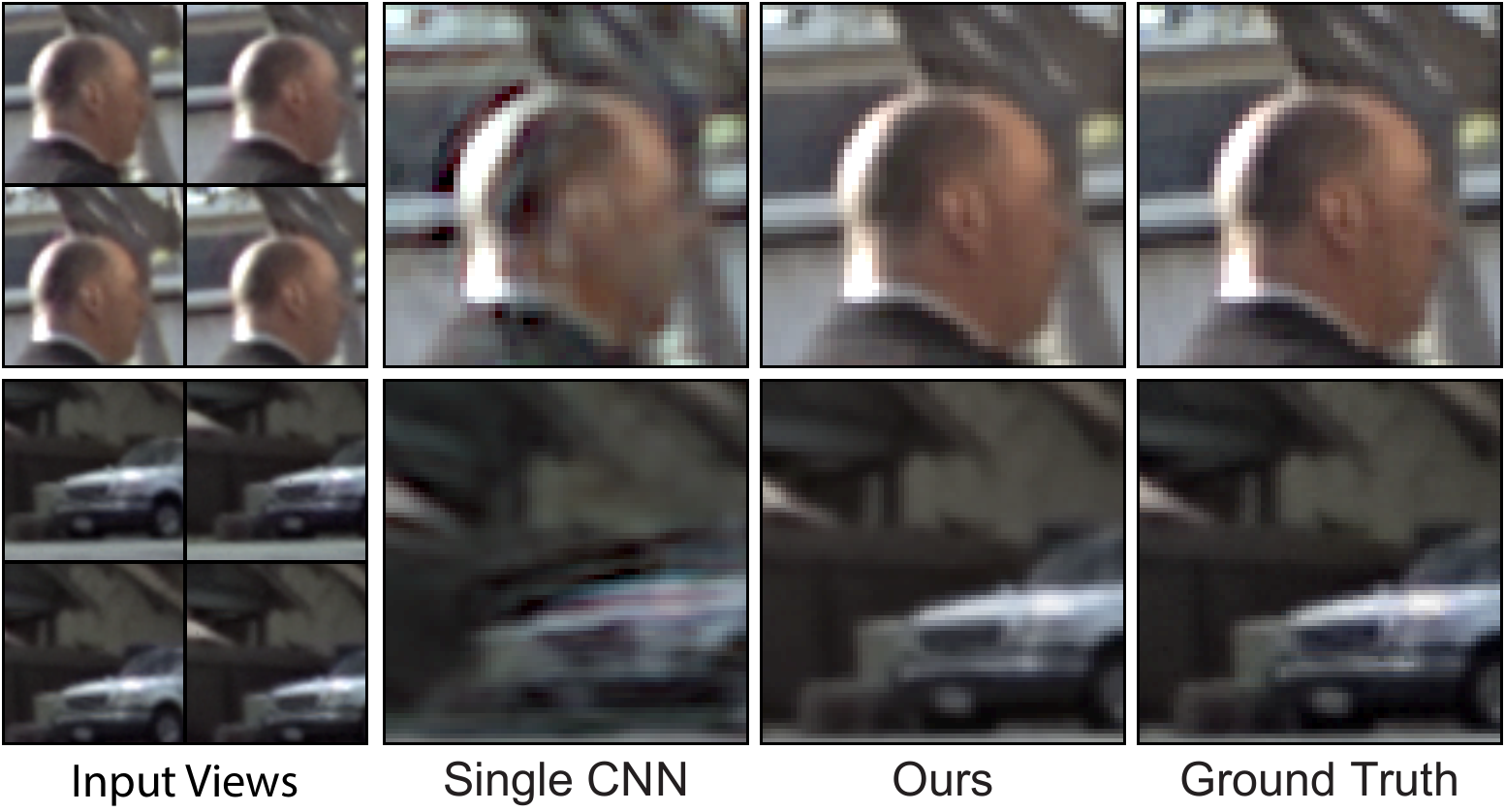}
\vspace{-0.2in}
\caption{We compare our approach against single CNN architecture. We show an inset of the {\sc Flower 1} and {\sc Rock} scenes on the top and bottom rows, respectively. One CNN is not able to model the complex relationship between the input images and the novel view image, and thus, produces results with ghosting and other artifacts. In contrast, our system containing two sequential CNNs is able to properly model the relationship and produce high-quality results which are comparable to the ground truth. Comparison of all the synthesized views can be found in the supplementary video.}
\label{fig:Naive}
\vspace{-0.1in}
\end{figure}

\begin{figure*}[!ht]
\centering
\vspace{-0.1in}
\includegraphics[width=\linewidth]{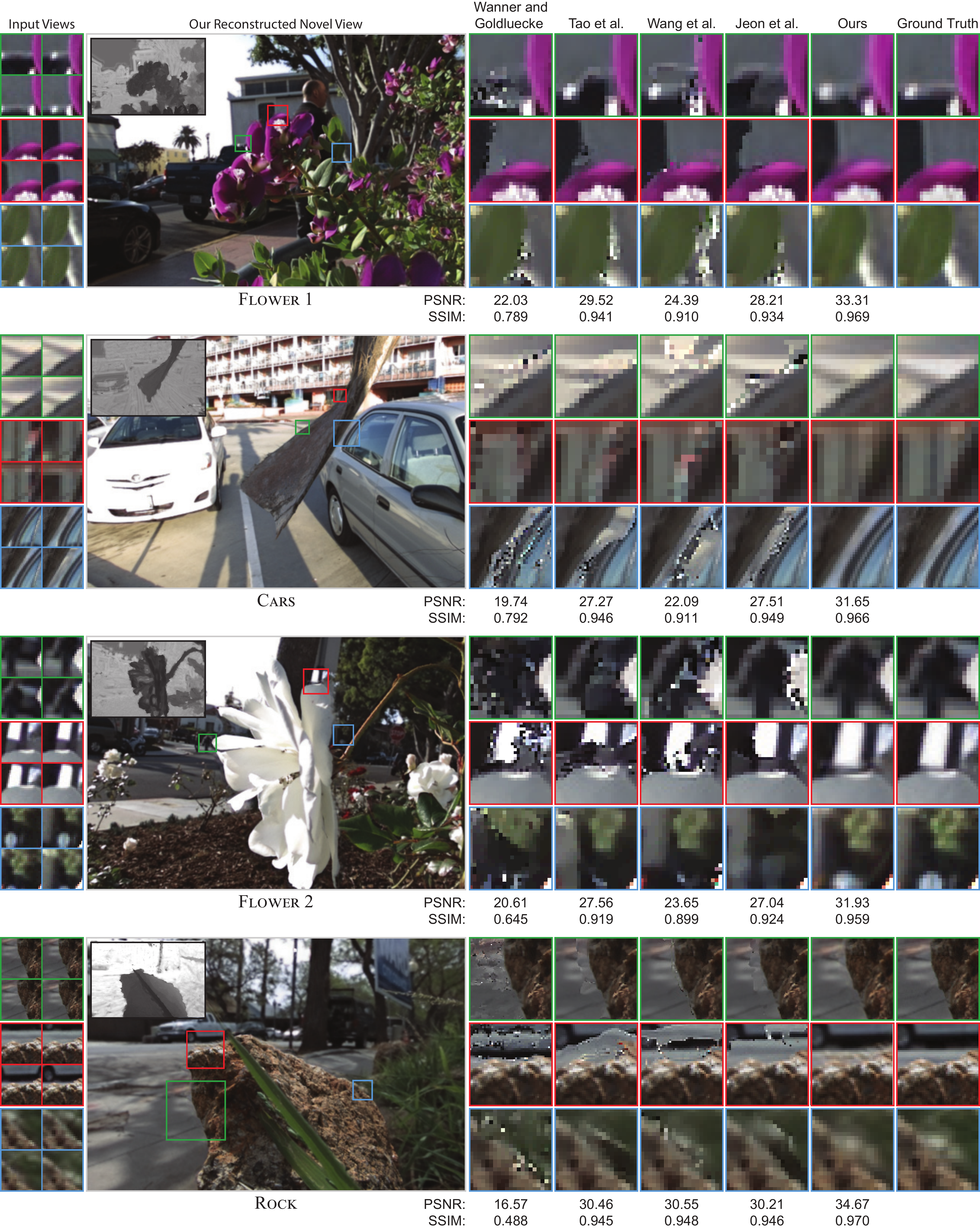}
\vspace{-0.1in}
\caption{We compare our approach against the method of Wanner and Goldluecke~[2014] using several state-of-the-art light field disparity estimation methods as its input. For each scene, we also show our estimated disparity, where regions with darker color are closer to the camera. The PSNR and structural similarity (SSIM) values are listed below each result. Note that some artifacts are hard to see in still images, and thus, we refer the readers to also view the supplementary video.}
\vspace{-0.2in}
\label{fig:MainComparison}
\end{figure*}

\begin{figure*}[t]
\centering
\includegraphics[width=\linewidth]{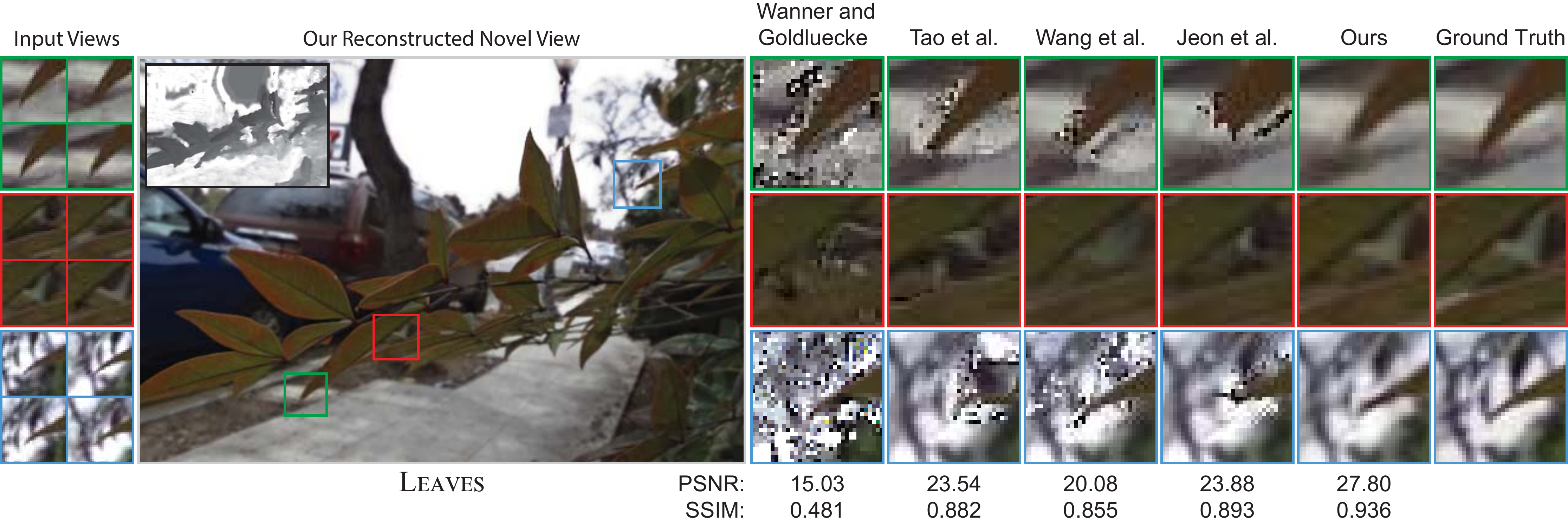}
\vspace{-0.2in}
\caption{Comparison of our approach against other methods on a challenging scene. The leaves have thin structures and the scene contains a significant number of occluded regions which makes the problem difficult. Our approach produces a reasonable result that is better than other methods. See the video comparison in our supplementary video materials.}
\label{fig:Challenging}
\vspace{-0.2in}
\end{figure*}

\section{Results}
\label{sec:Results}

We implemented our approach in MATLAB and used MatConvNet~\cite{Vedaldi15} for implementing our networks. All the results shown here are generated on light fields captured with a Lytro Illum camera. The angular resolution of the captured light fields is $8 \times 8$ from which we only use the four corner sub-aperture images as our input to generate the full light field. Note that our approach can generate any in-between views. However, we only generate the $8 \times 8$ views to be able to compare them against the ground truth images. Here, we only show one synthesized image (5, 5) for each scene, but videos showing all the views can be found in the supplementary video.

{\bf Comparison Against a Single Network} -- We begin by comparing to the result of modeling the process using a single CNN in Fig.~\ref{fig:Naive}. Here, the network directly models the relationship between the input images and the novel view (see Eq.~\ref{eq:Goal}). However, the relationship is complex and requires the network to often connect distant pixels, which makes the training difficult. As a result, when compared to our architecture containing two sequential CNNs, the result of the single CNN is blurry and contains artifacts. For example, the single CNN is not able to connect the pixels of the white truck in the input views, and thus, generates a result with ghosting artifacts.

\begin{table}
\begin{center}
%\begin{scriptsize}
\begin{tabular}{c | c | c | c | c | c}
 & Wanner & Tao & Wang & Jeon & Ours\\
 \hline
 \hline
 PSNR (dB) & 30.47 & 34.33 & 32.78 & 34.02 & 37.50\\
 SSIM & 0.861 & 0.954	& 0.946 & 0.952 & 0.970\\
\end{tabular}
%\end{scriptsize}
\vspace{-0.1in}
\caption{Comparison of our approach against other methods. The PSNR and SSIM values are averaged over 30 test scenes.} 
\label{tab:Comparison}
\vspace{-0.2in}
\end{center}
\end{table}

{\bf Comparison Against Other Approaches} -- Next, we compare our method against Wanner and Goldluecke's approach~\shortcite{Wanner14}. They first compute the disparity for each input view using an existing technique. They then use the disparities within an optimization framework to obtain the novel view by minimizing an objective function. We adopt several state-of-the-art light field disparity estimation methods to generate the disparities required for Wanner and Goldluecke's method. Specifically, we use the approaches by Wanner and Goldluecke~\shortcite{Wanner12}, Tao et al.~\shortcite{Tao13}, Wang et al.~\shortcite{Wang15}, and Jeon et al.~\shortcite{Jeon15}. We evaluate the results numerically, in terms of PSNR and structural similarity (SSIM)~\cite{Wang04}. SSIM produces a value between 0 and 1, where 1 indicates perfect perceptual quality with respect to the ground truth.

Table~\ref{tab:Comparison} shows the average PSNR and SSIM values for all the methods on 30 test scenes. To properly evaluate our system on challenging cases, we used images of foliage and flowers in about half of our test set. Note that we had completely separate training and test sets and none of the test scenes were part of the training set (see supplementary materials). As seen, our approach produces results that are significantly better than other methods. We show four of these scenes in Fig.~\ref{fig:MainComparison}. The {\sc Flower 1} scene demonstrates a flower in front of a truck, a building, and a tree (on the right). The flower and the leaves have complex structure which makes it hard for the other approaches to accurately estimate the disparity at the boundaries. Therefore, their results often contain artifacts around the occlusion boundaries. However, our approach produces a plausible result which is reasonably close to the ground truth image. Note, for example, that only our approach is able to properly reconstruct the truck's roof (green inset) and the highlight (blue inset).

Next, we examine the {\sc Cars} scene showing a tree branch in front of a street. Despite the simplicity of the scene, other approaches often are not able to accurately estimate the disparity around the boundaries of the branch from only four input images. Therefore, their result contains tearing artifacts which can be specifically seen in the blue inset. Moreover, the method of Wanner and Goldluecke~\shortcite{Wanner14}, which is used to synthesize the novel view, does not model the inaccuracies of consumer light field cameras which usually appear as discoloration in the results (see the colorful pixels in the red inset). Note that only our approach is able to reconstruct all the details around the occlusion boundaries such as the thin vertical line in the red inset.

The {\sc Flower 2} scene contains a flower with complex structure in front of a street. Our method produces a reasonable result that is better than other approaches. Note that only our method is able to faithfully reconstruct the challenging area between the flower stem and petal (blue inset). Finally, the {\sc Rock} scene is difficult for all the other approaches. They often are not able to accurately estimate the disparity around the boundaries of the rock which results in tearing artifacts. Meanwhile, we produce better results than the other methods relative to the ground truth. 

Overall, all the other approaches show tearing, ghosting, and other artifacts around the occlusion boundaries which are important areas for the view synthesis application. The main reason is that these methods are not specifically designed for this application, and thus, they often have inaccuracies around these boundaries. Moreover, Wanner and Goldluecke's approach~\shortcite{Wanner14}, which is used for generating the novel views, assumes the images are captured under ideal conditions, while this is not the case for consumer light field cameras. Our method, on the other hand, produces plausible results which are reasonably close to the ground truth. Numerically, our results are significantly better than the other approaches.

\begin{figure}[!ht]
\centering
\ifx\ShowTempImages\undefined
	\includegraphics[width=\linewidth]{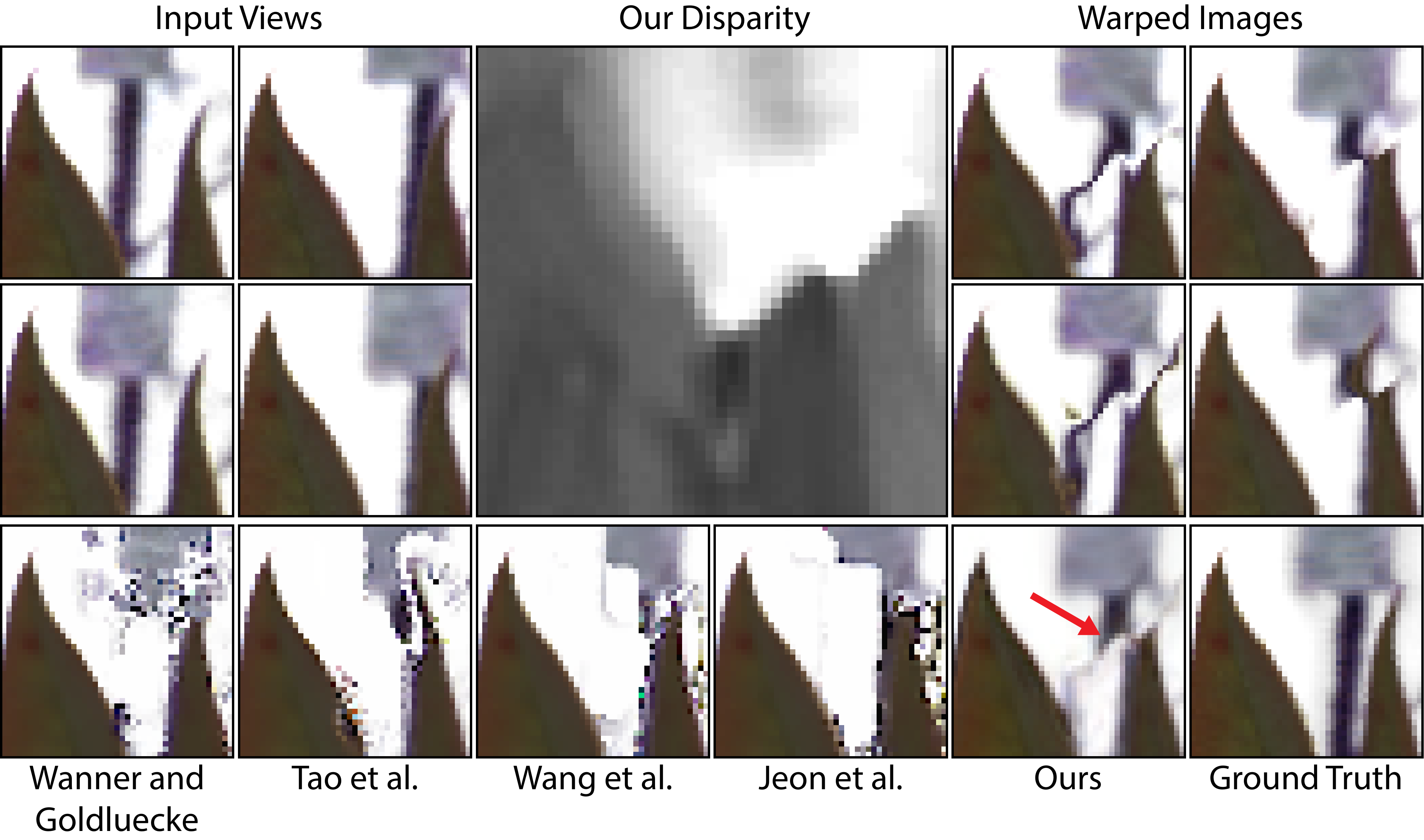}
\else
	\includegraphics[width=\linewidth]{Images/FailureCase_PlaceHolder}
\fi
\vspace{-0.25in}
\caption{\label{fig:Failure}
We show our result for an inset of the {\sc Leaves} scene. Our approach fails to correctly estimate the disparity in the area between the leaves. In this case, none of the warped images have the post in between the leaves which is visible in the ground truth image. Since our color predictor estimates the final image from these warped images, it fails to produce a high-quality image. However, our result is still more plausible than other methods.}
\vspace{-0.3in}
\end{figure}

We compare our method against other approaches on a challenging scene in Fig.~\ref{fig:Challenging}. This scene contains a significant number of occluded regions which are generally difficult for view synthesis. Therefore, even our approach fails to synthesize a high-quality image in the difficult regions (see Fig.~\ref{fig:Failure}). However, our result is reasonable overall and significantly better than all the other approaches. Note that the leaves have thin structure and only our approach is able to properly reconstruct them without introducing artifacts in the background (green and blue insets). 

\begin{figure}[t]
\centering
\includegraphics[width=\linewidth]{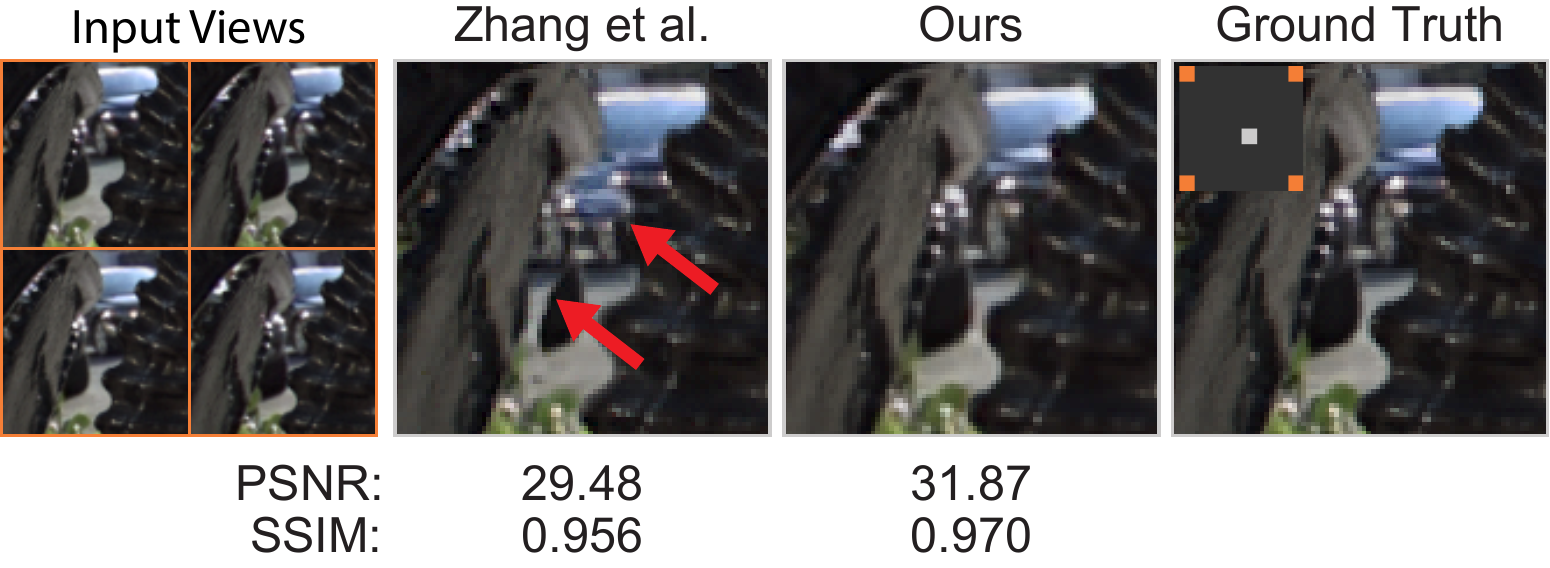}
\vspace{-0.25in}
\caption{Comparison of our approach against the recent method of Zhang et al.~\shortcite{Zhang16} on the {\sc Seahorse} scene. The inset shown here is a bigger version of the red inset in Fig.~\ref{fig:Motivation}.}
\label{fig:PlenoPatch}
\vspace{-0.15in}
\end{figure}

\edit{We also compare our approach against the recent method of Zhang et al.~\shortcite{Zhang16} in Fig.~\ref{fig:PlenoPatch}. Note that their approach needs some user interaction, while ours is fully automatic. Moreover, their method requires the center view, and thus, uses five input images (instead of four). Nevertheless, even with user interaction, their approach is not able to properly decompose the scene into different depth layers, resulting in tearing artifacts.}

{\bf Timing} -- Our method takes around 12.3 seconds to generate a novel view from four input images of resolution $541 \times 376$ on an Intel quad-core 3.4 GHz machine with 16 GB of memory and a GeForce \edit{GT} 730 GPU. Specifically, it takes 5.5 seconds to extract the features, 5.1 seconds to evaluate the disparity estimation network, 0.2 seconds for warping the four input images to the novel view, and 1.5 seconds to evaluate the color predictor network.

\begin{figure}[t]
\centering
\ifx\ShowTempImages\undefined
	\includegraphics[width=\linewidth]{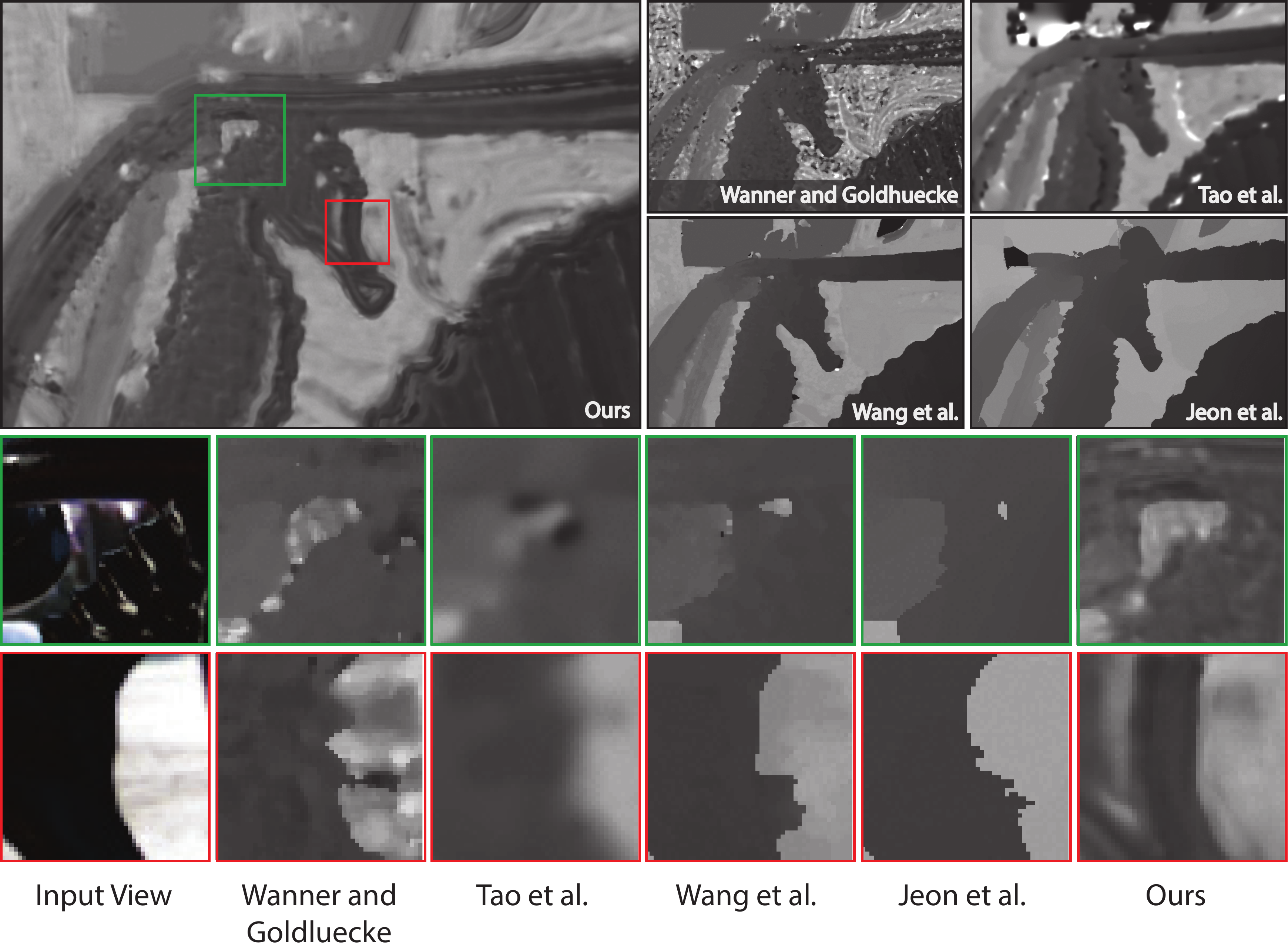}
\else
	\includegraphics[width=\linewidth]{Images/DepthComparison_PlaceHolder}
\fi
\vspace{-0.2in}
\caption{\label{fig:DepthComp}
We compare our estimated disparity with other approaches. The darker color indicates regions that are closer to the camera. Although some of the other techniques produce disparity with higher quality than ours, they are often inaccurate around the occlusion boundaries. Our approach produces a reasonable disparity and is more accurate in the occlusion boundaries which are important regions for view synthesis.}
\vspace{-0.2in}
\end{figure}

\begin{figure}[!ht]
\centering
\ifx\ShowTempImages\undefined
	\includegraphics[width=\linewidth]{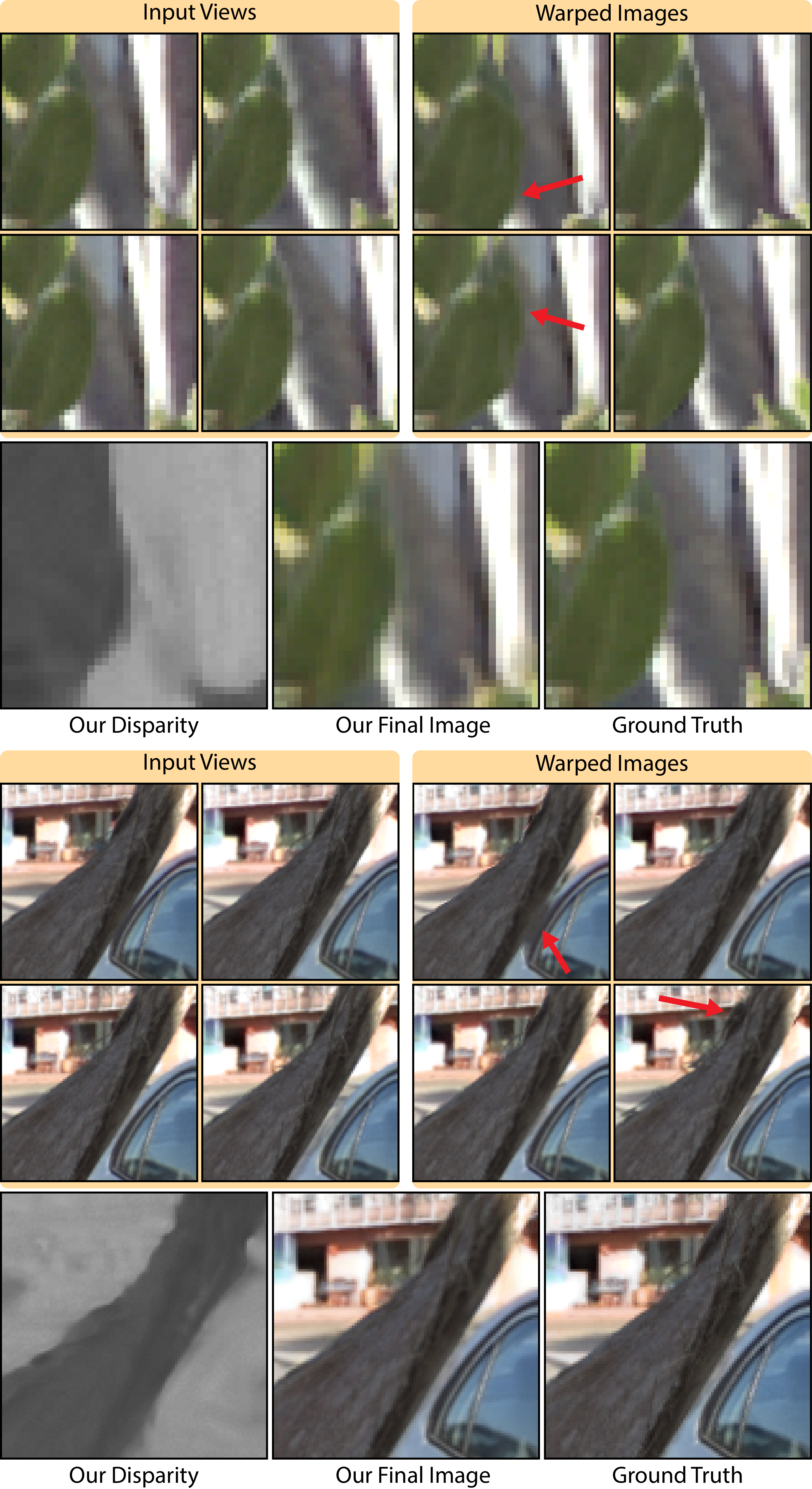}
\else
	\includegraphics[width=\linewidth]{Images/ColorPredictorEval_PlaceHolder}
\fi
\vspace{-0.2in}
\caption{\label{fig:ColEstEval}
We evaluate the effect of our color predictor network. On the top, we show an inset of the {\sc Flower 1} scene containing a leaf in the foreground covering part of a tree trunk in the background. On the bottom, we show an inset of the {\sc Cars} scene. Although our disparity captures the boundaries correctly, some of the warped images contain invalid information because of occlusion. For example, the leaf in the left input views on the top set blocks part of the tree trunk and the highlight which should be visible in the novel view. Therefore, as shown by the red arrows, the left warped images contain artifacts in these regions. Our color predictor network produces high-quality results by collecting appropriate information from all the warped images. 
}
\vspace{-0.3in}
\end{figure}

{\bf Analyzing Our System} -- We evaluate the effect of each component in our system. Figure~\ref{fig:DepthComp} compares our estimated disparity against other approaches for the {\sc Seahorse} scene (shown in Fig.~\ref{fig:Motivation}). Although the disparities produced by some of the other approaches have higher quality than ours, their disparity often has artifacts around the occlusion boundaries which are the most important regions for view synthesis. For example, these methods are not able to appropriately estimate the disparity of the background at the middle of the green inset, or the boundaries of the seahorse snout in the red inset. As a result, they often produce artifacts in these regions which can be seen in our supplementary video. 

Our method, on the other hand, can produce a reasonable disparity in these regions. Note that our method does not always produce accurate disparity. However, our inaccuracies usually happen in the regions that are not important for view synthesis. For example, part of the seahorse snout is incorrectly detected as background in our disparity (white region in left part of the red inset). However, this is a constant color region, and thus, this inaccuracy does not affect the quality of the synthesized results (see Fig.~\ref{fig:Motivation}). This is due to the fact that we train our disparity estimator network by directly minimizing the error between the synthesized and ground truth images. In the future, it would be interesting to combine our learning scheme with the ideas from the existing disparity estimation approaches to generate a more accurate disparity.

Next, we evaluate the effect of our color predictor network in Fig.~\ref{fig:ColEstEval}. Here, we show an inset of the {\sc Flower 1} and {\sc Cars} scenes on the top and the bottom, respectively. We use the estimated disparity at the novel view to warp all the input views to the novel view. Due to occlusion, these warped images often contain artifacts, as indicated by the red arrows. Our color predictor network properly detects these regions and produces a high-quality image by collecting appropriate information from the warped images.

{\bf Denoising Effect} -- Since we use all the input views to generate the novel views, our results are generally less noisy compared to the ground truth images. We refer the readers to our supplementary video to see this effect. This could potentially be useful for capturing light fields in low light conditions where noise is an issue.

{\bf Limitations} -- Our color predictor network generates the final image using the warped images. Therefore, in cases where none of the warped images contain valid information, our approach is not able to produce high-quality results. One of these cases is shown in Fig.~\ref{fig:Failure} for the {\sc Leaves} scene. Here, our approach fails to synthesize the post in between the leaves and produces tearing artifacts. However, our result is considerably better than other approaches.

\edit{Moreover, as shown in the supplementary video, our method can be used for extrapolation. However, since we specifically train our networks for interpolation, our extrapolation results have generally lower quality. Nevertheless, our method still produces better results than other approaches.}

\edit{Finally, although in this paper we focused on light fields obtained by consumer cameras, we believe a similar architecture can be adapted for unstructured light fields with larger disparities. However, as with any learning-based techniques, our system needs to be retrained to be able to properly work for these cases.}

\section{Conclusions and Future Work}
\label{sec:Conclusion}

We have presented a novel learning-based approach for synthesizing novel views from a sparse set of input views captured with a consumer light field camera. Our system consists of disparity and color estimator components which we model using two sequential convolutional neural networks. We show the result of our approach on a variety of scenes using only the four corner sub-aperture images captured with a Lytro Illum camera. Experimental results show that our method outperforms state-of-the-art approaches.

In the future, we would like to investigate the possibility of using our system for generating high dynamic range light fields from a set of views with different exposures. Moreover, it would be interesting to extend our system to work with any number of input views. We are also interested in improving the speed of our algorithm to possibly work at interactive rates or even real-time. Finally, there is potential to use our system along with light field compression schemes~\cite{Tong03,Girod03} to increase the compression ratio by, for example, generating the novel views from a sparse set and compressing the differences.

\section*{Acknowledgments}
\edit{We would like to gratefully thank Alexei Efros for valuable discussions. We also thank the Stanford Computational Imaging group for the light field dataset, some of which were used in our training set. Fang-Lue Zhang ran his algorithm on our scenes. This work was funded in part by a Berkeley Fellowship, ONR grant N00014152013, NSF grants 1451830 and 1617234, and the UC San Diego Center for Visual Computing, as well as support from Draper Lab, Nokia, and a Google Research Award.}

\begin{normalsize}
\bibliographystyle{acmsiggraph}
\bibliography{ViewSynthesis}
\end{normalsize}
\end{document}